\pdfoutput=1 
\documentclass[letterpaper, 10 pt, conference]{ieeeconf}  % Comment this line out if you need a4paper

\IEEEoverridecommandlockouts                              % This command is only needed if 
\usepackage{subfig}
\usepackage{hyperref}
\hypersetup{
    colorlinks=true,
    linkcolor=blue,
    filecolor=blue,      
    urlcolor=blue,
    pdftitle={Overleaf Example},
    pdfpagemode=FullScreen,
    }
\usepackage{times}
\usepackage{graphicx}
\usepackage{amsmath,amssymb,amsfonts}
\usepackage{wasysym}
\usepackage{comment}
\usepackage{multirow}
\usepackage{tikz}
\usepackage{listings}
\usepackage{pdfpages}
\usepackage{algorithm}
\usepackage{algpseudocode}

\usepackage{booktabs}

\newcommand\tomasys[1]{\emph{#1}}

\newcommand\pddl[1]{#1}

\title{\LARGE \bf
Runtime Architecture and Task Plan Co-Adaptation for Autonomous Robots with Metaplan}
\author{Jeroen M. Zwanepol$^{*}$, Gustavo Rezende Silva$^{*}$ and Carlos Hernández Corbato$^{*}$% <-this % stops a space
\thanks{$^{*}$ Delft University of Technology, Delft, The Netherlands
        {\tt\small j.m.zwanepol@student.tudelft.nl; \{g.rezendesilva, c.h.corbato\}@tudelft.nl;} 
        }%
\thanks{This work was supported by the European Union’s Horizon 2020 Framework Programme through the MSCA network REMARO (Grant Agreement No 956200).}% <-this % stops a space
}

\begin{document}

\maketitle
\thispagestyle{empty}
\pagestyle{empty}

\begin{abstract}
    Autonomous robots need to be able to handle uncertainties when deployed in the real world. For the robot to be able to robustly work in such an environment, it needs to be able to adapt both its architecture as well as its task plan. Architecture adaptation and task plan adaptation are mutually dependent, and therefore require the system to apply runtime architecture and task plan co-adaptation. This work presents Metaplan, which makes use of models of the robot and its environment, together with a PDDL planner to apply runtime architecture and task plan co-adaptation. Metaplan is designed to be easily reusable across different domains. Metaplan is shown to successfully perform runtime architecture and task plan co-adaptation with a self-adaptive unmanned underwater vehicle exemplar, and its reusability is demonstrated by applying it to an unmanned ground vehicle.
\end{abstract}
\section{Introduction}\label{sec:intro}

While operating, autonomous robots are subject to uncertainties, both from internal system factors (e.g., sensor failures) and external environment factors (e.g., unexpected obstacles). To overcome these uncertainties, robots can be designed as self-adaptive systems~\cite{weyns2020introduction}, enabling them to adapt at runtime their architecture (i.e., deactivating or activating software components and changing their parameters) and their mission execution (i.e., change the task being performed).
However, as stated by Cámara et al.~\cite{camara2020software} the mutual dependencies between architecture and task plan adaptation pose a significant challenge when determining how to apply both adaptations in a coordinated manner. 
Thus, this paper focuses on designing a solution for runtime architecture and task plan co-adaptation (RATPA).

In the context of self-adaptive robotic systems, several works have addressed either task adaptation, architecture adaptation, or both adaptations independently \cite{aguado2021functional, molina2021behavior, cheng2020ac, valner2022temoto, cicada}, but only few works apply RATPA \cite{camara2020software, braberman2015morph}. Both these works define the co-adaptation problem and propose solutions for it, however, the solutions presented are not general and reusable for different robotic applications.  

This paper proposes Metaplan, a planning based approach that uses off-the shelf general reasoners and knowledge about the robot’s architecture and mission to drive runtime architecture and task plan co-adaptation. 
Metaplan combines Metacontrol \cite{aguado2021functional}, a knowledge-based framework for architecture self-adaptation, with a Planning Domain Definition Language (PDDL)-based planner~\cite{Ghallab98} used for simultaneous architecture and task planning, ensuring the selection of a suitable configuration for each task. This results in a general and modular solution that is only dependent on: (1) a PDDL formulation of the robot's task planning problem, and (2) an architectural model of the robot conforming to Metacontrol's metamodel. This explicit dependency on models makes Metaplan reusable across different applications.

Metaplan is implemented as a ROS 2-based system leveraging MROS ~\cite{corbato2020mros, mros_package} and PlanSys2 ~\cite{martin2021plansys2, plansys_package}, a ROS implementation of Metacontrol and a package for handling plan generation and execution in ROS 2 respectively. It was evaluated with two different types of robots, an Unmanned Underwater Vehicle (UUV)~\cite{suave}, and an Unmanned Ground Vehicle (UGV)~\cite{camara2020software}. The experiments showed that Metaplan is able to perform RATPA in robotic systems, and that it is general and reusable across different applications with minimal overhead.

In summary, the contributions of this paper are:
\begin{itemize}
    \item A reusable approach for runtime architecture and task plan co-adaptation in robotic systems with minimal overhead;
    \item Experimental results demonstrating runtime architecture and task plan co-adaptation in the SUAVE exemplar;
\end{itemize}
\section{Related Works}\label{section:related_works}

In the context of self-adaptive robotic systems, there are multiple works that have addressed either architecture adaptation or task adaptation~\cite{aguado2021functional, molina2021behavior,carreno2021situation}, independent architecture and task adaptation~\cite{valner2022temoto, cicada}, or coordinated task and architecture adaptation~\cite{camara2020software, braberman2015morph}.  Table \ref{tab:related_works} presents an overview of the related works.

As a solution for architecture adaptation, Hernández et al.~\cite{hernandez2013model} proposed Metacontrol, a knowledge-based framework that incorporates systems with the ability to adapt their architecture at runtime.  
An important aspect of Metacontrol is that it uses 
knowledge of how the system is designed to reason when and how the system should adapt. 
This design choice increases Metacontrol's reusability, since it is only required to modify the knowledge base in order to apply it to different applications. Metacontrol was demonstrated 
with different robotics applications, such as mobile base navigation and underwater vehicles~\cite{aguado2021functional, hernandez2013model, hernandez2018self, hernandez2019meta, mros_darko}.

Purandare et al.~\cite{cicada} proposed the CICADA framework to enable Unmanned Aerial Vehicles (UAVs) to recover from misconfiguration-related flight instability. This is achieved by adapting the UAVs' flight controller configuration at runtime to a safe baseline configuration or by adapting the task it is performing. Although CICADA is able to perform both architecture and task adaptation, they are performed independently. In relation to reusability, CICADA was specifically designed to handle misconfiguration-related flight instability in UAVs, and does not generalize to other applications. 

Valner et al.~\cite{valner2022temoto} proposed the TeMoto software architecture to address the challenges of dynamic task and resource management. 
% Dynamic task management refers to the problem of being able to switch at runtime the robot's mission specification when requested by an operator, and configuring the robot properly for each of the mission's actions. 
% Note that this definition differs from the task adaptation definition used in this work, where task adaptation refers to adapting the tasks being performed due to changes in the state of the robot and environment. 
% Dynamic resource management relates to the issue of managing the resources required by the robot tasks. 
Similarly to CICADA, TeMoto is able to handle architecture and task adaptation independently, but not in a coordinated manner. In terms of reusability, TeMoto is task independent, but it requires the adaptation logic to be manually implemented for each managed resource.

RATPA was first addressed by Braberman et al.~\cite{braberman2015morph} with the MORPH reference architecture. 
% Their solution adopts a divide and conquer strategy, where the architecture is divided into two independent parts, one responsible for architectural concerns and another for task concerns. Both parts do not require explicit knowledge about each other to handle their own adaptation concern, and both parts negotiate to guarantee consistency in the selection of configurations and tasks.
Although MORPH is proposed as a solution for architecture and task plan co-adaptation, it is only demonstrated on a conceptual level. 
This makes it difficult to evaluate the feasibility of applying MORPH to robotic systems at runtime, and to compare it to other methods for robotics adaptation.

Cámara et al.~\cite{camara2020software} present a framework for architecture and task co-adaptation based on probabilistic model checking. 
Their approach consists of using 
% Alloy~\cite{jackson2002alloy} to define the robot's architecture and constraints, and to generate legal configurations of the robot given its current state. Furthermore, it uses 
% PRISM~\cite{kwiatkowska2011prism} 
an explicit model of the robot with a probabilistic model checker to synthesize all possible reconfiguration plans, including their associated cost, and to find the most suitable task plan for the robot, including the required reconfiguration plans. 
% An interesting aspect of their solution is that it guarantees the selection of an optimal configuration as the probabilistic model checker explores the entire solution space. However, exploring the entire solution space in large configuration spaces may take a significant amount of time, which could result in being unfeasible to adapt the system at runtime.
Some limitations of their approach are that task planning does not take into account that an action might become unavailable when there is no architectural configuration available to solve it. Furthermore, their task planning solution is specific for navigation on a graph-like map, thus applying it 
% their framework 
to a new task planning scenario requires extensive changes.

In the current state of the art, as far as the authors are aware, the only solution available for architecture and task plan co-adaptation is the one proposed by Cámara et al.~\cite{camara2020software}. However, their solution is not general enough to be applied to different applications without modifications. Therefore, this work aims to fill this gap by proposing Metaplan.

\begin{table}[]
\centering
\caption{Related adaptation frameworks}\label{tab:related_works}
\resizebox{\columnwidth}{!}{%
\begin{tabular}{@{}cccccc@{}}
\toprule
\multirow{2}{*}{Approach} & \multicolumn{3}{c}{Adaptation type}                                                                   & \multirow{2}{*}{\begin{tabular}[c]{@{}c@{}}Robotics\\ Middleware\end{tabular}} & \multirow{2}{*}{Reusable} \\ \cmidrule(lr){2-4}
                          & Architecture & Task & RATPA &                                                                                &                           \\ \midrule
Metacontrol~\cite{hernandez2013model}               & Yes           & X    & X                                                                              & ROS, ROS 2                                                                     & Yes                       \\
CICADA~\cite{cicada}                    & Yes           & Yes  & X                                                                              & ROS                                                                            & X                         \\
TeMoto~\cite{valner2022temoto}                    & Yes           & Yes  & X                                                                              & ROS                                                                            & Yes                       \\
MORPH~\cite{braberman2015morph}                     & Yes           & Yes  & Yes                                                                            & N/A                                                                            & X                         \\
Cámara et al.~\cite{camara2020software}             & Yes           & Yes  & Yes                                                                            & ROS                                                                            & X                         \\
\textbf{Metaplan}                  & \textbf{Yes}           & \textbf{Yes}  & \textbf{Yes}                                                                            & \textbf{ROS 2}                                                                          & \textbf{Yes}                       \\ \bottomrule
\end{tabular}%
}
\end{table}

\section{Metacontrol Background}\label{sec:metacontrol}
Metacontrol adheres to the principle of separation of concerns for self-adaptive systems into a managed subsystem and a managing subsystem \cite{weyns2020introduction}. Where the managed subsystem is responsible for domain concerns, in this case, the robotic application, and the managing subsystem is responsible for adapting the managed subsystem. Its reasoning cycle follows the MAPE-K loop \cite{horn2001autonomic}. It monitors the relevant parameters of the managed subsystem and the environment, analyzes whether the managed subsystem configuration fulfills its requirements, reconfigures when necessary.
All these steps are performed in interaction with its knowledge base (KB).  

\begin{figure}
    \centering
    \includegraphics[width=0.85\linewidth]{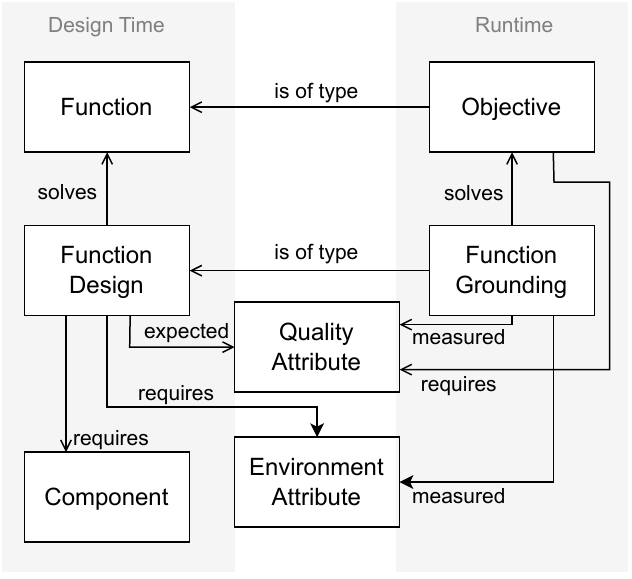}
    % \caption{A simplified representation of the TOMASys metamodel}
    \caption{TOMASys metamodel}
    \label{fig:tomasys}
\end{figure}

Metacontrol's KB conforms to the Teleological and Ontological Model for Autonomous Systems (TOMASys) metamodel \cite{hernandez2013model}. In short, TOMASys captures engineering knowledge of how the system is designed, its possible configuration alternatives, and how attributes of the managed system or environment relate to the selection of a configuration. A simplified version of TOMASys with the required concepts for this work can be seen in \autoref{fig:tomasys}.
The design time elements are specified by the designer of the adaptation logic, and they are static throughout the system operation. The runtime elements are dynamically specified by the system during its operation.
Throughout this work, TOMASys elements will be written with \tomasys{this font}.

The design time elements of TOMASys are: \tomasys{function}, \tomasys{function design}, \tomasys{component}, \tomasys{qualitty attribute}, and \tomasys{environment attribute}. A \tomasys{function} 
represents a functionality of the managed subsystem, e.g., navigation. 
A \tomasys{function design} 
is a particular solution for a \tomasys{function}, in other words, each \tomasys{function design} represents a possible configuration
that can be selected by Metacontrol, e.g., a particular navigation algorithm.  A \tomasys{component} 
represents a hardware or software component of the system that is required by a \tomasys{function design}, e.g., a sonar. A \tomasys{quality attribute} 
represents an attribute of the system, e.g., battery level. And a \tomasys{environment attribute} 
represents an attribute of the environment, e.g., water visibility. Specific \tomasys{quality attribute} or \tomasys{environment attribute} values can be set as requirements for \tomasys{function designs}, e.g., a certain navigation algorithm can only be used if the battery level is above $50\%$. 

The runtime elements of TOMASys are: \tomasys{objective} and \tomasys{function grounding}.
An \tomasys{objective} represents the \tomasys{functions} of the system that are currently required, e.g., while performing a navigation task the robot has an \tomasys{objective} of the navigation \tomasys{function} type. A \tomasys{function grounding} represents the current \tomasys{function design} that is deployed to solve an \tomasys{objective}.

In addition to the architectural model, the KB contains rules that are reasoned over at runtime to evaluate the status of the system and decide when adaptation is required, i.e., a rule indicating that a \tomasys{function grounding} is in error when the measured value of a \tomasys{quality attribute} violates its requirement.

The architectural model is implemented using the Ontology Web Language (OWL) \cite{Antoniou2004}. The rules to evaluate configurations are expressed with the Semantic Web Rule Language (SWRL)~\cite{swrl2004semantic}, and the rules are verified with the Pellet reasoner~\cite{sirin2007pellet}.
\section{Metaplan}

Metaplan is designed to perform the role of a managing subsystem. Metaplan assumes that the managed subsystem is component-based and task oriented. The former means that the manged subsystem is composed of multiple loosely coupled software components that can be activated or deactivated at runtime. The latter means that the managing subsystem is designed to handle tasks, which are realized by a sequence of actions. The managed subsystem is designed to accomplish these actions by activating the corresponding components.

\begin{figure*}[!h]
    \centering
    \includegraphics[width=0.67\textwidth]{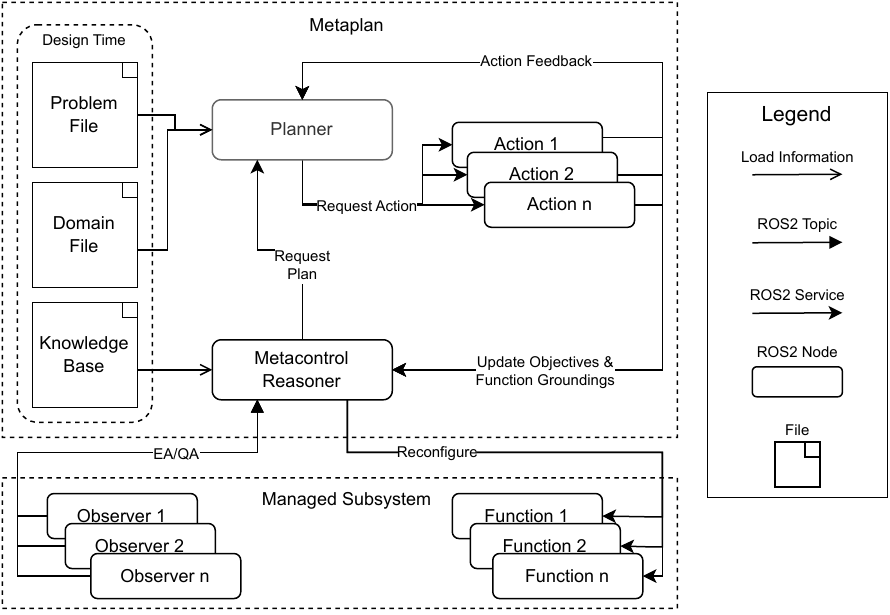}
    \caption{System Architecture of Metaplan. The system is split into a section containing the managed subsystem and a section containing the managing subsystem.}
    \label{fig:system}
\end{figure*}

% HERE
The architecture of Metaplan is depicted in \autoref{fig:system}. Its main components are the Planner and the Metacontrol Reasoner. The Metacontrol Reasoner is responsible for analyzing runtime information to decide when the robot needs to adapt, which configurations are currently available, and executing the reconfigurations selected by the Planner by activating and deactivating Function nodes of the managed subsystem. The Planner is responsible for planning the actions of the robot and its respective required configurations, and informing Metacontrol about the selected reconfiguration plans. The planner is based on PDDL, and its planning problem is defined in the Domain and Problem files.

%The Metacontrol Reasoner is responsible for analyzing runtime information to decide when the robot needs to adapt, which configurations are currently available, and for executing the reconfigurations selected by the Planner by activating and deactivating Function nodes of the managed subsystem.  
% For this, at runtime, Metacontrol reasons over the KB.

To reach its goal state, Metaplan generates a task plan consisting of a sequence of actions, and selects the required configuration to accomplish each action. These actions correspond to one or more \tomasys{functions} of the robot expressed in the KB, and the configurations for each action correspond to the selected \tomasys{function designs}. 
When an action finishes execution, the next action of the task plan starts execution.

Whenever a \tomasys{function design} becomes (un)available or an \tomasys{objective} is in error, Metaplan generates a new task plan. This allows for Metaplan to adapt the system configuration using the updated available \tomasys{function designs}, and could adapt the task plan by generating a different plan, thereby achieving runtime architecture and task plan co-adaptation.

The remainder of this section details Metaplan's architecture, the design time activities required to configure Metaplan for a robotic scenario, and its runtime behavior.

% \subsection{Metaplan architecture}

% The KB, the Domain File, and the Problem File specify the information required to configure Metaplan for a robotic mission with adaptation logic.

% The Metacontrol Reasoner takes information on \tomasys{quality} and \tomasys{environment attributes} obtained from Observer nodes and analyzes the KB to see which configurations are available to the robot and whether adaptation is required. When adaptation is required, the Metacontrol Reasoner requests a plan from the Planner. The Planner generates a plan with information from the Domain File and Problem File, together with the updated information on available configurations from the Metacontrol Reasoner. An action of the task plan is selected, which upon start of execution updates the objectives of the Metacontrol Reasoner. New configurations are then requested to the managed subsystem, corresponding with the new objectives. Alternatively, new configurations are requested by the Planner when an action finishes executing and a new action is requested.

\subsection{Design time activities}\label{sec:design}

To apply Metaplan to a specific application, it is necessary to create a KB containing an architectural model of the robot, and define the robot task planning problem with PDDL in a Domain and Problem file.

\subsubsection{Knowledge Base}
The KB captures information about the robot's architecture and its variants, it is modeled conforming to the TOMASys metamodel (see \autoref{sec:metacontrol}). The KB is reusable for different applications within the same domain, that is, for the same robot with the same architecture in a similar environment.

\subsubsection{Domain File}

The Domain File defines which actions are available, and their preconditions and effects. The Domain File contains a generic description of the actions the robotic system is able to perform, which conform to the \tomasys{functions} defined in the KB. This means that the Domain File can be used for different uses within the same domain, i.e., a Domain File containing a description of a mobile robot used in a hospital environment can be used for multiple missions for that same robot in the same environment.

% An important design requirement for the actions defined in the Domain File is the addition of a precondition describing which \tomasys{functions} from the KB are required for their execution. Moreover, there needs to be another precondition specifying that a \tomasys{function design} needs to be available in order to fulfil the required \tomasys{function(s)}. This allows the PDDL planner to select an action to be executed, together with the best available configuration for it, i.e., the best \tomasys{function design} available to solve the required \tomasys{functions}. 

An important design requirement for each action is that it needs a precondition stating which \tomasys{functions} it requires, and that there need to be \tomasys{function designs} available to solve these \tomasys{functions}. The PDDL formulation of these requirements can be seen in Listing \ref{lst:pddl_ac}. Assume action1 is one of the actions the system can perform. The action would first need to be linked to an action object (line 1). This action object requires a \tomasys{function} (line 2). This \tomasys{function} is solved by a \tomasys{function design} (line 3), which is available (line 4).  This allows the PDDL planner to select an action to be executed, together with a suitable configuration.

\begin{lstlisting}[
  caption={PDDL Required Action Definitions},
  label={lst:pddl_ac}]
1:    (at start (action1 ?a))
2:    (at start (a_req_f ?a ?f))
3:    (at start (fd_solve_f ?fd ?f))
4:    (at start (fd_available ?fd))
\end{lstlisting}

\subsubsection{Problem File}

The Problem File defines the initial state and goal state of the planning problem. This definition is specific to the mission of the robot and only applies within this context.

% An important design requirement for the Problem File is that each action needs to be linked to the \tomasys{functions} they require in the KB. For this a predicate is specified to link an object to a specific action. This object is then linked to the \tomasys{functions} it requires.

An important design requirement for Metaplan is information on which actions require which \tomasys{functions}. The PDDL formulation of this requirement can be seen in Listing \ref{lst:pddl_prob}. The object specifying action1 (line 1) and which \tomasys{functions} it requires (line 2) need to be created by the user.

\begin{lstlisting}[
  caption={PDDL Required Problem Definitions},
  label={lst:pddl_prob}]
1:    (action1 a1)
2:    (a_req_f a1 f1)
\end{lstlisting}

% This way a \tomasys{function design} solving the \tomasys{functions} of an action can be selected together with that action in order to generate a plan.

\subsection{Runtime behavior}

This section explains the runtime behavior of the components of Metaplan depicted in Figure \ref{fig:system}. Upon startup, the information from the KB, the Domain File, and the Problem File is loaded by the corresponding nodes.

The Metaplan loop starts by monitoring the \tomasys{environment} and \tomasys{quality attributes}. This information is passed to the Metacontrol reasoner. 
The Metacontrol Reasoner loop can be seen in Algorithm \ref{alg:metacontrol}. First (line 1), the Metacontrol Reasoner updates the Knowledge Base with information on the \tomasys{environment} and \tomasys{quality attributes} received from the Observer nodes. Next (line 2), the KB is analyzed with automatic reasoning using the KB rules to find which \tomasys{function designs} are available and whether any \tomasys{objectives} are in error. If the available \tomasys{function designs} have changed since the last iteration of the loop or there are \tomasys{objectives} in error, then the new \tomasys{function designs} are sent to the Planner and a new plan is requested (lines 3-5). Lastly (line 6), the old \tomasys{function designs} are updated with the new \tomasys{function designs} for the next iteration.

A plan request triggers the planner to generate a new plan. Algorithm \ref{alg:metacontrol} gives a step-by-step overview of how this is done. First (line 1), the Problem File (PF) is updated with the \tomasys{function designs} that are available. Afterwards (line 2), the Planner generates an action plan using the Problem File and Domain File (DF) consisting of a list of actions in sequence. These actions are then executed one by one, until the whole action plan is finished (lines 3-5).

An action is requested together with information on which \tomasys{function designs} (i.e., which configuration) should be used to solve the \tomasys{functions} required by the action. At the start of execution of an action, the \tomasys{objectives} and \tomasys{function groundings} of the Metacontrol Reasoner are updated. This, in turn, triggers the managed subsystem to use the Function nodes corresponding to the required \tomasys{functions} and to use the updated \tomasys{function groundings}.

\begin{algorithm}
    \caption{Metacontrol Reasoner}
    \label{alg:metacontrol}
    \begin{algorithmic}[1]
        \State $KB \gets$ Update($KB, QA, EA$)
        \State $FD_{new}, O_{error} \gets$ Analyze($KB$)
        \If{$FD_{new} \ne FD_{old}$ $\textbf{or}$ $O_{error} \ne \emptyset$}
            \State RequestPlan($FD_{new}$)
        \EndIf
        \State $FD_{old} \gets FD_{new}$
    \end{algorithmic}
\end{algorithm}

\begin{algorithm}
    \caption{Planner}
    \label{alg:pddl}
    \begin{algorithmic}[1]
        \State $PF$ $\gets$ Update($PF, FD_{new}$)
        \State $A \gets$ Plan($DF, PF$)
        \For{$a$ in $ A$}
            \State RequestAction($a$)
        \EndFor
    \end{algorithmic}
\end{algorithm}

\section{Evaluation}
Two experiments were performed to demonstrate the reusability and feasibility of Metaplan. The first experiment consisted of using Metaplan to design a solution for the UGV scenario described by Cámara et al.~\cite{camara2020software}. The goal of this experiment was to demonstrate the reusability and minimal overhead required for Metaplan following the design activities presented in \autoref{sec:design}. The second experiment consisted of applying Metaplan to a simulated unmanned underwater vehicle performing a pipeline inspection mission. The goal of this experiment was to demonstrate the feasibility to use Metaplan with robots. 

\subsection{UGV scenario}

In this scenario, the UGV has to navigate from an initial position to a goal position in the least amount of time and as safely and efficiently as possible. During its operation, the robot encounters obstacles and changing lighting conditions, to which it needs to adapt its architecture and task plan. 

\begin{figure}[]
    \centering
    \includegraphics[width=\columnwidth]{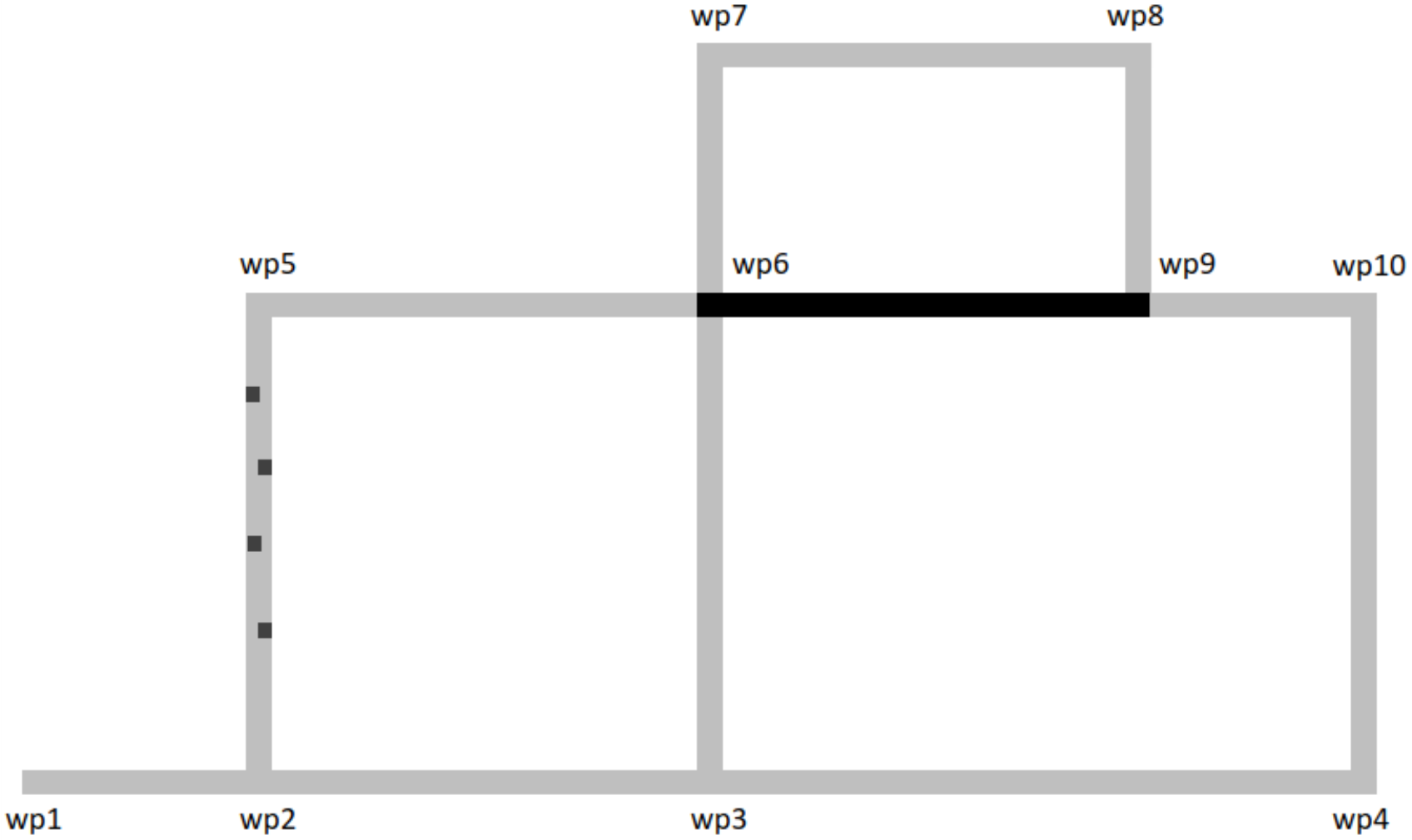}
    \caption{Adapted map of the UGV scenario}
    \label{fig:map}
\end{figure}

The robot has three distinct algorithms for localization: AMCL, MRPT, and Aruco. Each of these algorithms require input from different sensors, AMCL and MRPT require input from a lidar or a kinect, and the Aruco algorithm requires camera input. However, the camera can only be used in low light conditions in combination with a flashlight.

The robot navigates in a graph-like environment, with vertices representing intersections, and edges representing corridors. The robot knows the layout of the environment, and which corridors are dark or contain obstacles.
A simplified version of the map of the environment can be seen in \autoref{fig:map}. Note that the corridor between wp6 and wp9 is dark, and the corridor between wp2 and wp5 contains obstacles. At runtime, the robot needs to decide which paths to take and which configuration to use for each path.

The setup for the original experiment is not publicly available, thus this work is only simulated at the task and configuration level.

\subsubsection{Design time activities}
To apply Metaplan to this case, the design activities described in \autoref{sec:design} are followed, and they are described below.

\paragraph{Knowledge Base}
The only task the UGV is able to perform is navigating from point A to B, and the only variability available is which localization algorithm in combination with which sensors to use. Thus, only a localization \tomasys{function} called  \tomasys{f\_localization} is included in the ontology.
Each combination of the available localization algorithms with the available components is represented as a distinct \tomasys{function design} requiring specific \tomasys{components}, as can be seen in the first two columns of \autoref{tbl:fds_mm}.

Each \tomasys{function design} has an expected battery usage that is represented as a \tomasys{quality attribute} named \tomasys{QA\_battery\_usage}. Furthermore, each \tomasys{function design} requires a minimum level of the \tomasys{environment attributes} that represent how safe a path is and the lighting condition of the environment, respectively, \tomasys{EA\_safety} and \tomasys{EA\_light}. The specific values for the \tomasys{quality} and \tomasys{environment attributes} associated to each \tomasys{function design} can be seen in the last three columns of \autoref{tbl:fds_mm}.

\begin{table}[]
\centering
\caption{UGV Knowledge Base}
\label{tbl:fds_mm}
\resizebox{\columnwidth}{!}{%
\begin{tabular}{|c|c|c|cc|}
\hline
\multirow{2}{*}{\textbf{Function Design}} &
  \multirow{2}{*}{\textbf{\begin{tabular}[c]{@{}c@{}}Required \\ Components\end{tabular}}} &
  \textbf{QA} &
  \multicolumn{2}{c|}{\textbf{EA}} \\ \cline{3-5} 
 &
   &
  \textbf{\begin{tabular}[c]{@{}c@{}}Expected\\ Battery \\ Usage\end{tabular}} &
  \multicolumn{1}{c|}{\textbf{\begin{tabular}[c]{@{}c@{}}Required \\ Safety\end{tabular}}} &
  \textbf{\begin{tabular}[c]{@{}c@{}}Required\\ Light\end{tabular}} \\ \hline
fd\_AMCL\_lidar  & c\_lidar  & 4 & \multicolumn{1}{c|}{0.4} & 0 \\ \hline
fd\_AMCL\_kinect & c\_kinect & 2 & \multicolumn{1}{c|}{1.0} & 0 \\ \hline
fd\_MRPT\_lidar  & c\_lidar  & 6 & \multicolumn{1}{c|}{0.3} & 0 \\ \hline
fd\_MRPT\_kinect & c\_kinect & 4 & \multicolumn{1}{c|}{0.9} & 0 \\ \hline
fd\_aruco        & c\_camera & 7 & \multicolumn{1}{c|}{0.7} & 0 \\ \hline
fd\_aruco\_with\_light &
  \begin{tabular}[c]{@{}c@{}}c\_camera \& \\ c\_flashlight\end{tabular} &
  10 &
  \multicolumn{1}{c|}{0.7} &
  1 \\ \hline
\end{tabular}%
}
\end{table}

\paragraph{Domain File}
Three distinct move \pddl{actions} are defined, one for each type of corridor, e.g., \pddl{move}, \pddl{move\_with\_obstacle}, and \pddl{move\_dark}. 

\paragraph{Problem File}
The Problem File defines the \pddl{initial state} and \pddl{goal state} of the robot. It also contains information about all corridors between two waypoints and the type of each corridor, e.g., dark corridor or corridor with obstacles. Furthermore, the problem file also contains information on which \tomasys{functions} the \pddl{move actions} require, the current location of the robot, the battery level of the robot, and the goal location of the robot.

\subsubsection{Results}

The robot Metaplan updates the availability of the \tomasys{function designs} 
for each corridor comparing the \tomasys{EA\_light} and \tomasys{EA\_safety} levels specified in the KB (\autoref{tbl:fds_mm}) with the required levels corresponding to the environment map (\autoref{fig:map}).
The measured \tomasys{EA\_light} is 0 for the dark corridors and 1 for the other corridors, and the measured \tomasys{EA\_safety} is 0.8 for the corridors with obstacles and 0 for the other corridors. Note that the measured safety indicates the level of safety required to pass through that corridor, so Metaplan needs to select a \tomasys{function design} with an \tomasys{EA\_safety} of more than 0.8 to pass through those corridors.

\begin{figure}
    \centering
    \includegraphics[width=0.8\linewidth]{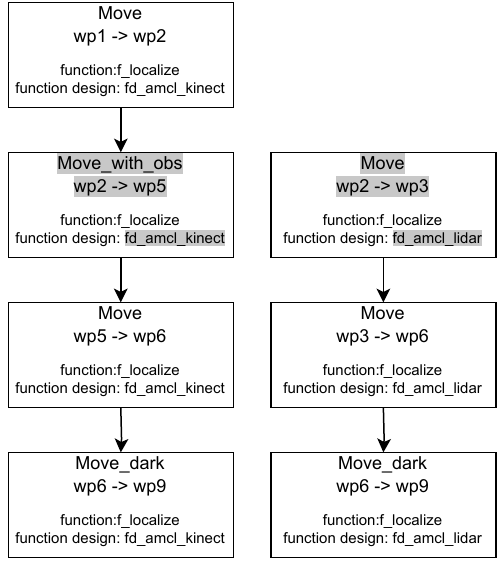}
    \caption{Initial plan (left) and adapted plan after kinect failure at wp2 (right). Move action, route, and function design have changed as a result.}
    \label{fig:plan_mm}
\end{figure}

For this experiment, the battery level is initially set to 100\% and the robot has to navigate from wp1 to wp9. The first plan generated by the robot is presented on the left side of \autoref{fig:plan_mm}. The robot selects \tomasys{fd\_amcl\_kinect}, since enough battery is available to select this \tomasys{function design}.
% , and the \tomasys{function designs} do not influence the duration of an action. 
Afterward, the robot encounters a corridor with obstacles and does not need to adapt its configuration, since \tomasys{fd\_amcl\_kinect} meets the required safety levels. 

When the robot reaches wp2, \tomasys{fd\_amcl\_kinect} and \tomasys{fd\_mrpt\_kinect} are set to be unavailable to simulate a failure of \tomasys{c\_kinect}. This requires the robot to adapt its task plan, since no \tomasys{function designs} are available for navigating through an obstacle corridor. Furthermore, the robot is also required to adapt its architecture, since \tomasys{fd\_amcl\_kinect} is unavailable. This results in the task plan presented on the right side of \autoref{fig:plan_mm}. Metaplan successfully finds a new plan avoiding the obstacle corridor and passing through wp3 instead, and adapts its configuration to \tomasys{fd\_amcl\_lidar}. After this, the robot can fulfil the rest of the mission without needing replanning.

The reusability of Metaplan can be demonstrated with the following modifications of the UGV scenario. 

\paragraph{Modification 1} Suppose that the UGV's initial position, goal position, or the map layout is changed. In comparison to the original scenario, only the PDDL problem file needs to be modified by changing the variables corresponding to the initial position, goal position, or map layout.

% \paragraph{Modification 2} Suppose that the move action can only start after a predetermined time of the day. In comparison to the original scenario, only the PDDL domain and problem file need to be modified by adding a precondition to the move action in the domain file, and adding the current time in the problem file.

\paragraph{Modification 2} Suppose that the move action in a corridor obstacle is altered to use more battery. In comparison to the original scenario, only the PDDL domain file needs to be modified by changing the values of the required battery level precondition and the battery consumption effect in the move action for the corridor with obstacles.

\paragraph{Modification 3} Suppose that the implementation of the MRPT localization algorithm is improved, thus its required safety levels can be changed. In comparison to the original scenario, only the KB needs to be modified by changing the values of the required \tomasys{EA\_safety} for \tomasys{fd\_mrpt\_kinect} and \tomasys{fd\_mrpt\_lidar}.

\subsection{UUV pipeline inspection}
This scenario consists of a UUV that is deployed to inspect a pipeline
as fast as possible, while ensuring it has enough battery available to complete the inspection. 
The UUV has a set of actions available with which it can achieve its mission. These actions are search pipeline, follow pipeline, and recharge. If at any time the battery falls below a critical level, the UUV needs to go to the charging station and recharge. Each action requires a different set of \tomasys{functions} and therefore different configurations of the UUV. 

The UUV needs to be able to handle two types of uncertainties: changing water visibility and a sudden drop in battery level. 
This requires the UUV to adapt both its architecture, to use less battery, and its task plan, to go to the charging station before fulfilling the rest of the mission.

The UUV is able to configure its speed to low, medium, or high. A high speed allows the UUV to reach its goal state faster, however, it consumes more energy.
The robot is also able to adjust the height at which it searches for the pipeline. There are three different heights the UUV can search at, namely low, medium, and high. A high search height allows the robot to detect the pipeline faster, however the UUV will not be able to see the pipeline if the water visibility is low. 

% \subsubsection{Design time activities}
% The KB includes the \tomasys{functions} required to perform the search pipeline, follow pipeline, and recharge actions. The \tomasys{function designs} corresponding to the different speed and search height configurations. A \tomasys{quality attribute} representing the UUV's battery level, and \tomasys{enviroment attribute} representing the water visibility. The Domain File contains a description of each action, and the Problem File defines the initial and goal states of the robot.

\subsubsection{Results}
\begin{figure}
    \centering
    \includegraphics[width=0.8\linewidth]{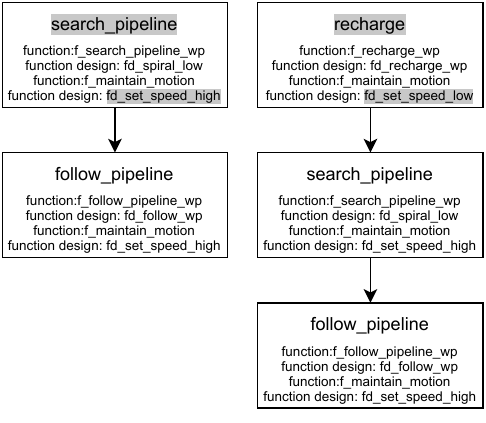}
    \caption{Initial plan (left) and adapted plan after sudden battery level drop (right). Action and function design have changed as a result.}
    \label{fig:plan_uuv}
\end{figure}

At the beginning of execution of the UUV's mission, an initial plan specifying a sequence of actions and required configurations was generated, which can be seen on the left side of \autoref{fig:plan_uuv}. The initial plan requires the UUV to first search for the pipeline and, when found, follow the pipeline. When a simulated sudden drop in battery level is inserted, the UUV is required to adapt and generates a new plan, which can be seen on the right side of \autoref{fig:plan_uuv}. The UUV changes its configuration by adjusting its speed to low and changes its plan by first going to the recharge station. 

This experiment was successfully executed in simulation using the SUAVE exemplar~\cite{suave}, demonstrating the feasibility of applying Metplan at runtime to a ROS 2-based robotic systems. A video of the simulation showcasing the RATPA behavior can be seen \hyperlink{https://youtu.be/Zi9dy0OUNUE}{here}.

\section{Conclusions and Future Works}
This work presented Metaplan as a reusable solution for architecture and task plan co-adaptation. Metaplan uses a PDDL-based planner for task and configuration planning, and Metacontrol to capture, analyze, and manage architectural knowledge. This resulted in a general and reusable framework since it only depends on a PPDL formulation of the robot's task planning problem, and in an architectural model of the robot conforming to the TOMASys metamodel, as detailed in \autoref{sec:design}. The experiments performed provided further evidence to the reusability claim by applying Metaplan to two distinct robotic applications, and demonstrated the feasibility of applying it to robotic systems at runtime.

Future works include: evaluating Metaplan's time performance and how it scales in more complex scenarios; capturing the dependency between tasks and functionalities in TOMASys instead of in the PDDL formulation, to make it more general and reusable.

\clearpage
% BIBLIOGRAPHY:
% use {unsrt}:
\bibliographystyle{ieeetr}
\bibliography{references}

% The following lines are necessary for showing the appendices correctly, do not change!

\end{document}